\documentclass{article} % For LaTeX2e
\usepackage{main,times}
\usepackage{graphicx}
\usepackage{subcaption}
\usepackage{hyperref}
\usepackage{url}

\title{Learning Invariances for Policy \\ Generalization}

\author{Remi Tachet des Combes, Philip Bachman \& Harm van Seijen \\
Microsoft Research Montreal\\
Montreal, QC H3A 3H3, CANADA \\
\texttt{\{retachet,phbachma,havansei\}@microsoft.com}
}

\begin{document}

\maketitle

\begin{abstract}
While recent progress has spawned very powerful machine learning systems, those agents remain extremely specialized and fail to transfer the knowledge they gain to similar yet unseen tasks. In this paper, we study a simple reinforcement learning problem and focus on learning policies that encode the proper invariances for generalization to different settings. We evaluate three potential methods for policy generalization: data augmentation, meta-learning and adversarial training. We find our data augmentation method to be effective, and study the potential of meta-learning and adversarial learning as alternative task-agnostic approaches.

%By augmenting the state space of the underlying MDP, we explicitly expose the agent to the type of invariances it requires to generalize properly. In the meta-learning setting, we ensure through an additional loss term that improvement on one training task leads to improvement on the others. Finally, our adversarial method aims at masking task-specific information from the feature space of the neural network in order to learn the proper problem invariants.
%While recent progress has allowed the training of very powerful machine learning systems, those agents remain extremely specialized and fail to transfer properly to unseen similar tasks. We consider that question in the reinforcement learning framework where we study a very simple toy problem on which standard RL algorithms do not generalize. By augmenting the state of the underlying MDP, we explicitly expose the agent to the type of invariances it requires to generalize properly and show the potential of data augmentation for reinforcement learning. We also discuss other methods aimed at implicitly learning invariances. We discuss two other methods aimed at implicitly learning invariances: a meta-learning approach aligning the gradient updates computed on different tasks and an adversarial method masking task-specific information in the feature space of the neural network.

\end{abstract}

\section{Introduction}

Deep Reinforcement Learning has produced impressive results in the recent past, allowing machines to master Go \citep{go}, beat Ms. PacMan \citep{pacman} or achieve complex robotic locomotion \citep{robots}. The algorithms used to achieve those feats are fairly generic, but their outcome is extremely specialized and cannot be used to play any other game: the methods developed for AlphaGo can be used on chess or shogi, but AlphaGo itself cannot play those games. The knowledge those systems gain does not transfer to other scenarios. Designing training algorithms and artificial agents that perform well out-of-the-box on tasks not encountered before is referred to as \textit{zero-shot learning} \citep{oh,darla} or \textit{domain generalization} \citep{da}, and is paramount to solving general intelligence. 

In this paper, we focus on an instance of the domain generalization problem commonly met in video games: controlling an avatar and making it jump over an obstacle. Variations of this objective (or tasks) are obtained by changing the position of the obstacle. Humans will only need a few tries on a few tasks to estimate where the avatar needs to jump and will easily infer the strategy to follow when the obstacle is somewhere else on the screen. We show that the same cannot be said of a machine trained from scratch on the same few tasks with classic reinforcement learning techniques. Because of the limited number of tasks at hand, there are many ways to solve them while depending heavily on task-specific features and not on the generalizing ones that carry over between tasks. We call learning the features that truly matter for a given problem \textit{learning invariances} and discuss alternatives to accomplish it. Our ultimate goal is the open problem of training an agent on a set of tasks as small as possible and having it generalize well to unseen obstacle positions.

%It will become clear very quickly that the quantity allowing to generalize to any task is the relative distance between the controlled object and the obstacle. a machine trained from scratch on a few tasks with classic reinforcement learning techniques might not generalize to unseen obstacle positions

%While seemingly trivial, we designed this problem to study policy generalization in a controlled environment where the dynamics are simple and yet generalization difficult for machines. However, it is far from obvious that training a neural network with classic reinforcement learning techniques will produce any representation of that relative distance. We aim here at studying how well standard algorithms generalize on the problem and the limitations they face, and eventually to propose a method to overcome those limitations. Our ultimate goal is to train an agent on a set of tasks as small as possible and have it generalize well to unseen combinations obstacle/floor height.

In the following, we study the effectiveness of three methods to learn invariances for policy generalization: data augmentation, meta-learning and adversarial training. After a brief overview of related work, we describe the tasks we are trying to solve and the details of the methods we evaluated. Finally, we discuss their results on policy generalization.

%Those systems perform super-humanly on the tasks they were trained on, they however utterly fail when exposed to another type of problem. The reinforcement learning training procedures used in the papers mentioned above are fairly generic, but an agent able to solve the Atari game Breakout cannot be used to beat Ms. PacMan, AlphaGoZero cannot play chess, an agent trained to direct a robot arm cannot steer a self-driving vehicle... 

%In the spirit of \cite{da}, we study an instance of the domain generalization problem. Let us consider the seemingly trivial task of controlling an avatar and making it jump over an obstacle. After a few tries on a few different positions of the obstacle, any human will estimate the position where the avatar needs to jump and easily infer the strategy to follow for any other obstacle position. On the other hand, deep model-free reinforcement learning techniques, exposed to the same limited number of obstacle positions, will perfectly solve those few tasks, but will undoubtedly fail to learn the invariances necessary to generalize to unseen situations.

\section{Related Work}

\textbf{Zero-shot learning} is a topic of great interest these days. Among others, \cite{oh} manually enforce analogies (resp. dissimilarities) between similar (resp. different) variables to reach better generalization on an 3-D labyrinth task. \cite{darla} apply $\beta$-VAE to extract independent latent variables from a scene and apply q-learning on those features. It is worth noting that independent features, while undoubtedly useful, are not sufficient on their own to guarantee generalization as shall be seen on our toy problem. Additionally, it is notoriously difficult to apply q-learning to a small set of features, e.g. the RAM of an Atari game\footnote{A difficulty we also encountered during certain experiments.} \citep{atariRAM}.

\textbf{Data augmentation} is a very common technique to induce invariances in supervised learning. On image recognition tasks, it usually involves randomly cropping and rotating the images of the data set, and results in agents with better generalization capabilities \citep{dataaugmentation}. In reinforcement learning, data augmentation has scarcely been applied. \cite{go} use the symmetries of the Go board to generate more games and arguably ensure the encoding of those symmetries in AlphaGo's policy and evaluation networks. \cite{oneshotopenai} pretrain a visual module on simulated objects with a variety of color, backgrounds, and textures in order to make the learnt representation independent from those attributes. That training is however done prior to the RL phase.

%That module's output is then used to train a block stacking robot. Data augmentation requires knowing beforehand the type of invariances required. It effectively entails trading supervision (through the choice of modifications applied to the data) for generalization, which limits its significance. 

\textbf{Meta-learning} has been the focus of much attention recently. \cite{maml} apply it to find model parameters such that a small number of gradient steps on a new task will produce good generalization performance on it. \cite{da} adapt the method to zero-shot learning via a procedure aimed at aligning the gradient updates for two arbitrary training tasks and show some generalization improvement on Cart-Pole and Mountain Car. Compared to data augmentation which requires transforming images in somewhat sophisticated manners, the only supervision meta-learning requires is a task identifier, making it a broader technique. 

In the wake of \cite{gan}, much research has been directed towards \textbf{adversarial training}. The original idea is to simultaneously train a generator that captures the data distribution, and a discriminator that estimates whether a sample is real or generated. \cite{domain} apply adversarial training to domain adaptation and produce shift-invariant classifiers. \cite{fader} extend the discriminator's duty to separate attributes (e.g. age, gender or glasses) from salient information in the latent space. In \cite{conf/nips/XieDDHN17}, the authors learn representations invariant to a \textit{specific factor or trait of data}. 

\section{Experiment and Methods}

\begin{figure}
\centering
    \begin{subfigure}{0.35\textwidth}
        \hspace*{-0.8cm}
        \includegraphics[width=\textwidth]{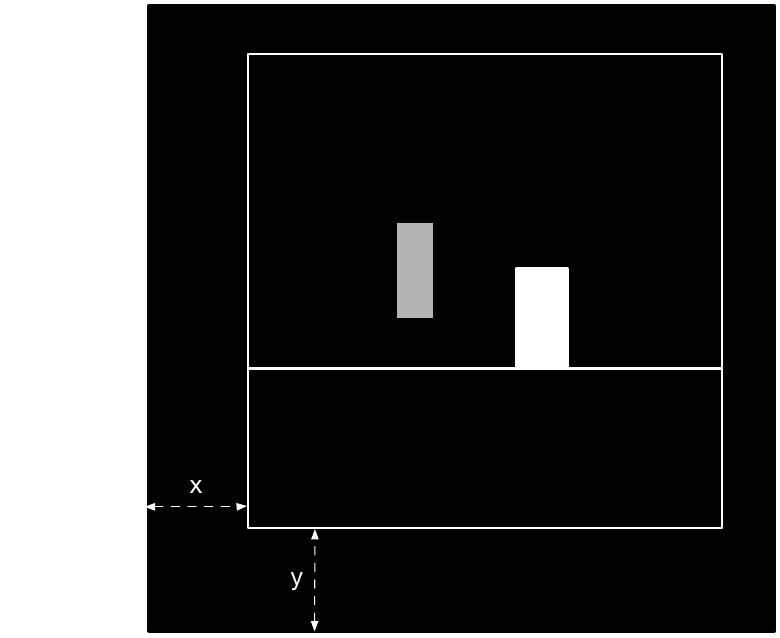}
    \end{subfigure} 
    \begin{subfigure}{0.19\textwidth}
        \hrule
        \includegraphics[width=\textwidth]{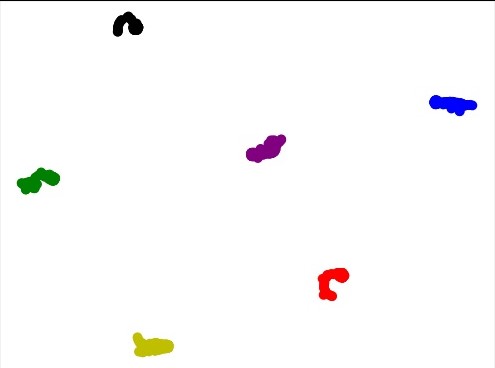}
        \hrule
        % \vspace*{-0.5cm}
        \includegraphics[width=\textwidth]{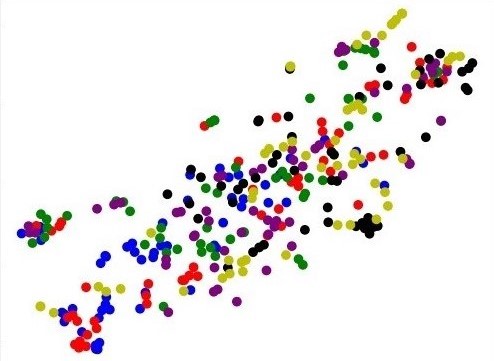}
        \hrule
    \end{subfigure} 
    \hspace*{0.3cm}  
    \begin{subfigure}{0.40\textwidth}
        \includegraphics[width=\textwidth]{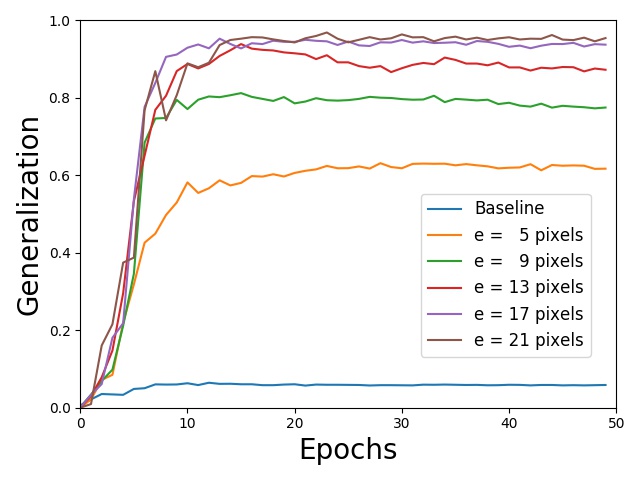}
    \end{subfigure}\hspace*{\fill}
% \centering
    % \begin{subfigure}[c]{0.4\textwidth}
    %     \includegraphics[width=\textwidth]{remi.pdf}
    % \end{subfigure}\hfill
    % \begin{subfigure}[c]{0.5\textwidth}
    %     \includegraphics[width=\textwidth]{task_5_positions_6.jpg}
    % \end{subfigure}\hspace*{\fill}
\caption{\textit{Left.} Embedded game screen. The grey rectangle is the controllable object or agent, the white one the obstacle. When data augmentation is performed, the game screen is embedded into a larger screen at a random position $(x,y)$ with $1 \leq x,y \leq e$. \textit{Centre.} t-SNE plots of hidden layer outputs over the course of an episode with (bottom) and without (top) adversarial training. Colors correspond to training tasks. \textit{Right.} Generalization performance of data augmentation for different values of $e$.}\label{game}
\end{figure}

\textbf{Setup.} We consider an extremely simple video game\footnote{The code for the game can be found at \url{https://github.com/Maluuba/jumping-task}.} consisting of a black background, a floor and two rectangles (Fig.\ref{game} \textit{Left}). The grey rectangle starts on the left of the screen and can be moved with two actions, "Right" and "Jump". The goal of this game is to reach the right of the screen while avoiding the white obstacle. There is only one specific distance (measured in number of pixels) to the obstacle where the agent has to chose the action "Jump" in order to pass over the obstacle. If jumping is chosen at any other point, the agent will inevitably crash into the obstacle. A reward of +1 is granted anytime the agent moves one pixel to the right (even in the air). The episode terminates if the agent reaches the right of the screen or touches the obstacle. We build a set of related tasks by varying two factors: the floor height and the position of the obstacle on the floor. The resulting set contains 1271 tasks. We use 6 of those for training and evaluate the generalization performance as the fraction of the remaining 1265 tasks the agent can solve.

% As there is one pixel position, and one only, where the agent has to jump to pass over the obstacle, solving an unseen task without a proper abstract representation of the problem is impossible. 

% \section{Methods}

% \section{Explicit Invariance Training Through State Augmentation}

\textbf{Explicit Invariance Learning Through State Augmentation.} We first attempt to guarantee generalization through data augmentation: we embed the original game screen at a random position in a larger black screen, and use that as input to classic RL algorithms. The larger screen has the shape of the game screen with $e$ pixels added in both dimensions (see Fig.\ref{game} \textit{Left} for details). The position of the game screen is kept constant during each episode.

%$e$ is the number of possible pixel positions (in x and y) of the game screen inside the augmented one. 

\textbf{Meta-learning Approach.} An interesting approach to zero-shot learning is to make sure that updates performed on one task are beneficial for solving other related tasks \citep{maml}. In our case, one can hope that if getting better on a given position makes the agent better on another one, its overall ability to jump over arbitrary obstacles will also improve. In the spirit of \cite{da}, this translates into the following optimization:
\begin{equation}
    \min_\theta L_1(\theta) + L_2(\theta - \alpha \nabla_\theta L_1)
\end{equation}
where $\theta$ represents the weights of the neural network, $L_1$ and $L_2$ the losses on two random tasks (in our case, two positions of the obstacle), and $\alpha$ the learning rate. $\theta - \alpha \nabla_\theta L_1$ represents the weights after an update made to improve on task 1, we want that update to also decrease the loss on task 2. A first order approximation of this minimization reduces it to aligning the gradient updates (in parameter space) obtained on different tasks, guiding the optimization to a more \textit{generalizing} part of the function space \citep{da}.

\textbf{Task Adversarial Learning.} Another interesting approach to this problem is to adversarially remove any task-specific information from the deep layers of the network. This is done through the addition of a discriminator trained to recognize the task instance from the output of a certain hidden layer. The RL agent then has two goals: solve the tasks and maximize the entropy of the discriminator's output. This approach is an RL equivalent to \cite{fader}'s Fader Networks and to \cite{conf/nips/XieDDHN17}'s framework.

\section{Results and Discussion}\label{results}

We trained an agent\footnote{The results shown are for the DQN network \citep{atari} but other architectures perform equivalently.} using standard Double-DQN \citep{doubleDQN} and A2C \citep{a2c} with and without data augmentation for a variety of $e$ values. The generalization performance of the agent is shown on Fig.\ref{game} \textit{Right}. Without data augmentation (blue curve) the agent solves all the training tasks, but only $36$ out of the $1265$ ($2.8\%$) testing tasks. With \textbf{data augmentation}, it is able to solve up to 1263 of the 1265 variations ($99.8\%$), confirming in a reinforcement learning setting that it is very helpful for generalization. Both DDQN and A2C produce the same patterns. Interestingly, the agent has also become able to jump over two consecutive obstacles, a situation never observed during training. \textbf{Meta-learning} and \textbf{adversarial training} on the other hand have so far failed to perform better than the baseline.
 
The one essential feature to learn in order to solve the jumping problem and generalize properly to any task is the \textit{relative distance} between the agent and the obstacle. In other words, the agent needs to learn translation invariance. Other features are irrelevant and using them might actually prevent generalization. For instance, the agent coordinates on the screen are sufficient to solve the problem in a variety of ways. One is to compute the difference between the two x-coordinates and learn which value should trigger a jump. Another is to discriminate between tasks using the coordinates of the obstacle, and then memorize for each task the absolute position where the agent needs to jump. The former will generalize while the latter will not\footnote{Learning independent features is thus in itself not enough to guarantee generalization \citep{darla}.}. Our results show that through \textbf{data augmentation}, the agent becomes able to learn translation invariance. That invariance however has to be enforced manually, effectively trading supervision for generalization. Contrary to data augmentation, \textbf{meta-learning} is an invariance-agnostic training procedure and as such bears more promise. Additional work is required to understand the dynamics of learning with the meta-learning term and hopefully get better generalization. With \textbf{adversarial training}, the agent is successful at performing the training tasks and at masking their specific information: a t-SNE plot of hidden layer outputs shows clear tasks' clusters without adversarial training but none with (Fig.1 \textit{Centre}). It appears though that the restriction provided by the adversary is not sufficient to generalize. Implicitly learning the proper invariance for generalization is still an open problem.

\bibliography{main}
\bibliographystyle{main}

\newpage

\setcounter{section}{0}
\renewcommand{\thesection}{\Alph{section}}
% \makeatletter
% \def\@seccntformat#1{%
%   \expandafter\ifx\csname c@#1\endcsname\c@section\else
%   \csname the#1\endcsname\quad
%   \fi}
% \makeatother

\section{Toy experiment}

The toy game we designed is in black and white with pixel values in the $[0,1]$ interval, and has the following characteristics:
\begin{itemize}
    \item The game screen is a black (pixel values $= 0$) $60 \times 60$ pixels square.
    \item The agent is a grey (pixel values $= 0.5$) $5 \times 10$ pixels rectangle.
    \item The obstacle is a white (pixel values $= 1$) $9 \times 10$ pixels rectangle.
    \item The floor is a white one-pixel thick horizontal line and is positioned at an arbitrary height between 0 and 40 pixels.
\end{itemize}
The agent starts on the floor, on the far-left of the screen. Two actions can then be performed, either \textit{Right}, which results in a single pixel motion to the right or \textit{Jump}. The jumping dynamics are simplistic: when the agent jumps, it moves in a straight oblique line to a height of $15$ pixels above the floor and then comes down in a symmetrical motion. When the agent is in the air, actions have no effects. Due to those dynamics, the obstacle's abscissa needs to be larger than $17$ for it to be avoidable by the agent (in which case, \textit{Jump} is the first action is needs to choose). The jumping parameters and agent sizes have been chosen so that the agent must pick the \textit{Jump} action at a unique position to succeed, anything else will result in the agent colliding with the obstacle. This was done to minimize the possibility of the agent succeeding by chance.

When performing data augmentation, we place the game screen at random within a larger 80x80 image. The number of possible positions is a parameter we explore (see Section \ref{results} and Figure \ref{game} \textit{Right}.). It varies from $1$ to $441$, with $e$ varying from $1$ to $21$.

Each time the agent 1 pixel to the right, it gets a $+1$ reward (this is also true in the air). The episode terminates if the agent touches the obstacle or reaches the right side of the screen. In each episode, the maximum score is $57$. During training, we used a discounting factor $\gamma = 0.99$.

To experiment with this toy experiment and its environment, please visit \url{https://github.com/Maluuba/jumping-task}.

\section{Training}

\subsection{Algorithms}

As baselines on our reinforcement learning task, we used using two standard instances of value-based and policy-based methods: DQN \citep{atari} and A2C \citep{a2c}. Using DDQN \citep{doubleDQN} did not lead to any particular improvement.

\subsection{Network Architectures}

\textbf{DQN and A2C} We experimented with different networks to assess their impact on generalization. The starting point was the original DQN architecture \citep{atari}: three convolutional layers with 32, 64 and 64 channels, filters of sizes 8, 4 and 3 and strides of 4, 2 and 1 followed by two fully-connected layers (the intermediate one of size 512). To take advantage of the translation invariance of convolutions, we tested using max-pooling at different depths of the network (to reduce the output dimension to the proper number) and no fully connected layers but it did not improve generalization. Similarly, having only one dense layer did not affect performance.

\textbf{Data Augmentation} We used the same architecture than for the baselines, the only difference is the input dimension. The agent performance on the original game and on an extended screen (with the game being embedded at a unique position) are identical.  

\textbf{Meta-learning} This approach simply involves a different loss, the architecture used was identical to the baseline one.

\textbf{Adversarial Training} The discriminator used is a fully connected network, we tested different number of hidden layers and neurons in those layers and found no significant impact. The results given in the main text were obtained with three hidden layers of 40 neurons. We also tried connecting the discriminator to the second and third hidden layer which did not affect the outcome of the experiments. The task information was properly masked in those activations, but the network did not learn the invariance to translation.

\subsection{Optimization}

Our networks were trained using a $0.001$ learning rate and the Adam optimizer \citep{adam} with the following parameters: $\beta_1=0.9$, $\beta_2=0.999$ and  $\epsilon=0.0001$. We manually explored part of the hyperparameters space and found no influence on the final outcome of the experiments.

\subsection{Methods}

\textbf{Meta-learning} To test the \textbf{meta-learning} approach, we augmented the classic DQN loss in the following way. Let $L_i$ be the standard DQN loss, computed on transitions from task $i$\footnote{$i$ stands for the $i$-th possible obstacle position.}, sampled from the experience replay. At each training step, we sampled two possible tasks $i$ and $j$ randomly and minimized:

\begin{equation}
    \min_\theta L_i(\theta) + \beta L_j(\theta - \alpha \nabla_\theta L_1) \nonumber
\end{equation}

where $\beta$ is an additional hyperparameter over which we swept manually.
We also attempted to train from a meta-train and a meta-test set (\textit{i.e.} sampling $i$ from a subset of $k$ possible tasks and $j$ from the rest of them), but it did not change the generalization performance of the model.  

\end{document}